%% file: ax_main.tex
\documentclass[10pt,twocolumn,letterpaper]{article}

\usepackage{iccv}
\usepackage{times}
\usepackage{epsfig}
\usepackage{graphicx}
\usepackage{amsmath}
\usepackage{amssymb}
\usepackage{balance}
\usepackage{kotex}
\usepackage{comment}

\usepackage{multirow} %
\usepackage{booktabs} %

\usepackage{bbm} %

\usepackage{cuted} %
\usepackage{capt-of} %
\usepackage{stfloats}

\usepackage[pagebackref=true,breaklinks=true,letterpaper=true,colorlinks,bookmarks=false]{hyperref} %

\iccvfinalcopy %

\ificcvfinal\fi

\input{misc/macro.tex}

\begin{document}

\title{Realistic Image Synthesis with Configurable 3D Scene Layouts}

% \author{
% Jaebong Jeong\\
% POSTECH\\
% Republic of Korea\\
% {\tt\small jbjeong@postech.ac.kr}
% \and
% Janghun Jo\\
% POSTECH\\
% Republic of Korea\\
% {\tt\small jhjo432@postech.ac.kr}
% \and
% Jingdong Wang \\
% Microsoft Research Asia\\
% Beijing, Chaina \\
% {\tt\small jingdw@microsoft.com}
% \and
% Sunghyun Cho\\
% POSTECH\\
% Republic of Korea\\
% {\tt\small s.cho@postech.ac.kr}
% \and
% Jaesik Park\\
% POSTECH\\
% Republic of Korea\\
% {\tt\small jaesik.park@postech.ac.kr}
% }

\author{
Jaebong Jeong$^1$ \and Janghun Jo$^1$ \and 
Jingdong Wang$^2$ \and Sunghyun Cho$^1$ 
\and Jaesik Park$^1$}
\date{%
    $^1$POSTECH \and
    $^2$Microsoft Research Asia\\[2ex]%
    % \today
}

\maketitle

\ificcvfinal\thispagestyle{empty}\fi

\begin{abstract}
\input{section/abstract.tex}
\end{abstract}

\input{section/intro.tex}

\input{section/related.tex}
\input{section/method.tex}

\input{section/experiment.tex}

\input{section/limitation.tex}
\input{section/conclusion.tex}

\clearpage

\begin{figure*}
% \onecolumn
% \begingroup
% \begin{center}
\centering
    {\Large Realistic Image Synthesis with Configurable 3D Scene Layouts\\ \it{Supplementary Material}}
% \end{center}
% \twocolumn
% \endgroup
\end{figure*}

\appendix
% \addcontentsline{toc}{section}{Appendices}
% \section{Supplementary Material}

\input{section/supplement.tex}

\clearpage

\balance
{\small
\bibliographystyle{ieee_fullname}
\bibliography{egbib}
}

\end{document}

%% file: misc/macro.tex
\usepackage{color}

\newcommand{\Fig}[1] {Figure \ref{fig:#1}}

\definecolor{purp}{rgb}{0.65, 0.16, 0.65}
\definecolor{greeen}{rgb}{0, 0.5, 0}

%% file: section/abstract.tex
Recent conditional image synthesis approaches provide high-quality synthesized images. However, it is still challenging to accurately adjust image contents such as the positions and orientations of objects, and synthesized images often have geometrically invalid contents. To provide users with rich controllability on synthesized images in the aspect of 3D geometry, we propose a novel approach to realistic-looking image synthesis based on a configurable 3D scene layout. Our approach takes a 3D scene with semantic class labels as input and trains a 3D scene painting network that synthesizes color values for the input 3D scene. With the trained painting network, realistic-looking images for the input 3D scene can be rendered and manipulated. To train the painting network without 3D color supervision, we exploit an off-the-shelf 2D semantic image synthesis method. In experiments, we show that our approach produces images with geometrically correct structures and supports geometric manipulation such as the change of the viewpoint and object poses as well as manipulation of the painting style.

%% file: section/intro.tex
\section{Introduction}

Conditional image synthesis aims to synthesize realistic image conditioning on user input to control the generation process.
Sparked by the recent success of deep learning-based generative models such as generative adversarial networks (GANs)~\cite{Goodfellow2014GenerativeAN},
conditional image synthesis has recently gained significant attention~\cite{Brock2019BIGGAN, Isola2017pix2pix, Kang2020ContraGAN, Karras2019StyleGAN, Karras2020StyleGANv2, Mansimov2016text2image, Miyato2018cGANsWP, Odena2017ConditionalIS, Park2019SemanticIS, Ramesh2021DALLE,  Sushko2020YouON_arxiv, Wang2018HighResolutionIS, Zhang2019SAGAN}.
For effective control of the generation process, various types of conditions have been explored such as class labels~\cite{Brock2019BIGGAN, Kang2020ContraGAN, Miyato2018cGANsWP, Odena2017ConditionalIS, Zhang2019SAGAN}, texts~\cite{Mansimov2016text2image,Ramesh2021DALLE}, style vectors~\cite{Karras2019StyleGAN,Karras2020StyleGANv2}, semantic layouts~\cite{Isola2017pix2pix, Park2019SemanticIS, Sushko2020YouON_arxiv, Wang2018HighResolutionIS}, and sketches~\cite{Isola2017pix2pix}.

Nevertheless, most recent conditional image synthesis approaches focus on providing a rough guide either on the global or local appearance of a synthesized image.
As a result, it is still challenging to accurately adjust image contents such as the positions and orientations of objects, and synthesized images often have geometrically invalid contents, e.g., disconnected legs of chairs.

\begin{figure}[t]
    \begin{center}
    \includegraphics[width=0.9\linewidth,page=1]{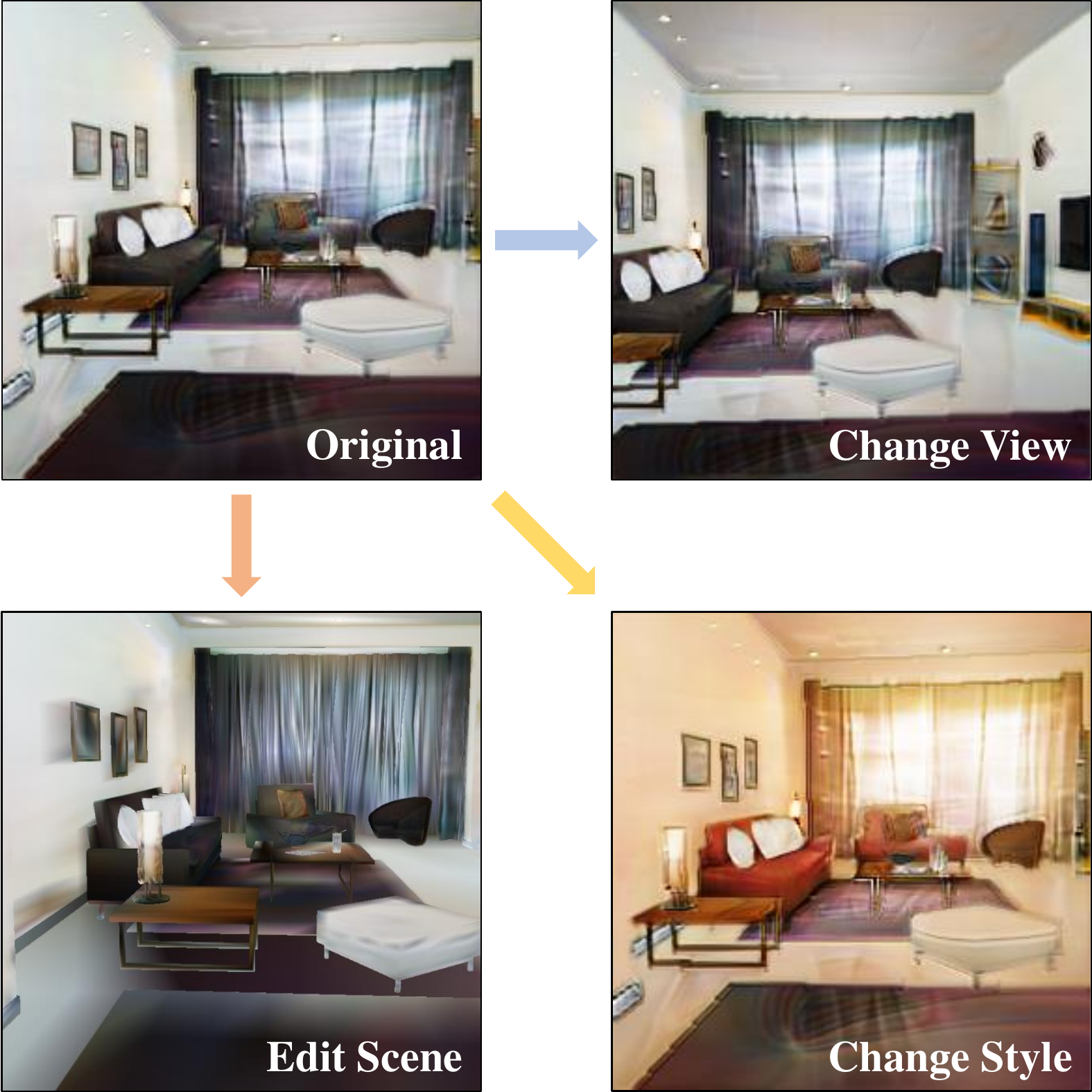}
    \end{center}
    \caption{Our method synthesizes images from 3D scenes and provides a way to change viewpoints, scene style manipulation, and scene editing. Our network learns to paint a scene with 3D geometry and semantic label map of the scene, and it does not require the color of 3D scenes as the supervision.
    }
    \label{fig:teaser}    
\end{figure}

To provide users with rich controllability on the image generation process in the aspect of 3D geometry,
in this paper, we propose a novel synthesis approach that we refer to as \emph{3D scene painting}.
Specifically, our approach takes a 3D scene consisting of multiple 3D models labeled with semantic class labels and their arrangement as input.
We consider a 3D scene as our input because it is highly controllable and can be easily designed with a preset of 3D models using a modeling tool, e.g., Blender~\cite{Blender}.
For an input 3D scene, we find a 3D scene painting function that synthesizes a color value for each point on each surface in the input 3D scene.
With the learned network, a user can render realistic-looking 2D images of the input 3D scene layout, and moreover, manipulate them based on 3D geometries, e.g., changing the viewpoint or object poses.
To the best of our knowledge, our approach is the first approach that synthesizes realistic images from a configurable 3D scene with multiple objects.

Our 3D scene painting approach offers several distinct advantages over the previous conditional image synthesis approaches~\cite{Isola2017pix2pix, Park2019SemanticIS, Sushko2020YouON_arxiv, Wang2018HighResolutionIS}.
First, it guarantees that synthesized images have geometrically valid structures.
Second, it allows a user detailed control of scene layouts as it takes a 3D scene as its input.
Third, once a 3D scene painting function is trained, a user can change the viewpoint or the positions of objects and produce images that are consistent with each other. 
Fourth, by conditioning the 3D scene painting function on a style vector, we can also produce images with the same 3D scene layout but different styles.
\Fig{teaser} illustrates operations that our approach supports.

For effective 3D scene painting, we model the painting function as a deep neural network (DNN) and train it for each input 3D scene.
To train a scene painting network to synthesize realistic color values for an input 3D scene, our framework utilizes a recent GAN-based semantic image synthesis approach that produces a realistic-looking image from a semantic segmentation map~\cite{Sushko2020YouON}.
For a given 3D scene layout, we synthesize multiple reference images from different viewpoints utilizing an image synthesis method.
As the images are generated independently, they do not have consistency across them.
We then train a 3D scene painting network using the reference images to aggregate information from them and synthesize consistent images regardless of the viewpoint.
To support multiple styles, we also adopt a style vector to the painting network.
Note that our painting network does not use 3D color supervision but uses a pre-trained semantic image synthesis method and 3D models with semantic class labels, which are easier to acquire.

In our experiments, we apply our approach to various indoor scenes.
Our qualitative and quantitative experimental results verify that our approach is highly controllable and produces high-quality realistic-looking images.

Our contributions can be summarized as follows:
\begin{itemize}
    \item We propose a novel approach to realistic-looking image synthesis based on a configurable 3D scene layout.
    \item We propose a 3D scene painting scheme that utilizes pre-trained semantic image synthesis methods without 3D color supervision.
    \item Our approach produces geometrically correct scene-level images, and supports geometric manipulations such as change of the viewpoint and object poses as well as the change of the style.
    \item We propose a new annotation to a 3D indoor dataset to generate configurable images with 3D components.
\end{itemize}

%% file: section/related.tex
\section{Related Work}

\paragraph{Image synthesis from 2D semantic label maps}
Since the emergence of GANs~\cite{Goodfellow2014GenerativeAN} and conditional GANs (cGANs)~\cite{Isola2017pix2pix}, there have been proposed a number of approaches that utilize 2D semantic label maps to control the image synthesis process.
Wang~\etal~\cite{Wang2018HighResolutionIS} proposed a coarse-to-fine generator and a multi-scale discriminator to achieve high-resolution synthesis from a 2D semantic label map.
Park~\etal~\cite{Park2019SemanticIS} proposed spatially-adaptive normalization (SPADE) layers.
Sch\"{o}nfeld~\etal~\cite{Sushko2020YouON} train a generator with a discriminator that is a semantic segmentation network.
Zhu~\etal~\cite{Zhu2020SMIS} proposed a group decreasing network based on group convolutions.
Ntavelis~\etal~\cite{Ntavelis2020SESAME} proposed a method for semantic editing of an image using semantic labels.
While these approaches show astonishing results, they do not consider 3D scene layouts.

\paragraph{3D-aware image synthesis}
There have been various attempts to incorporate 3D information for image synthesis such as novel view synthesis, texture reconstruction, and 3D-aware generative models.
Novel view synthesis approaches aim at generating images of novel viewpoints from a single input image or multiple images~\cite{Park2017TVSN,2019XuVIGAN,Tinghui2016AFlow}. However, they neither synthesize an entirely new image nor allow the adjustment of object poses.

Some attempts have also been made to handle textures of a single 3D object for image generation.
Grigorev~\etal~\cite{Grigorev2019CBI} and Huang~\etal~\cite{Huang2020AdvTexture} proposed image synthesis methods conditioned on the 3D pose of a single object.
Henderson~\etal~\cite{Henderson2020Leveraging2D} proposed a method for synthesizing a textured 3D mesh of a single object.
Martin-Brualla~\etal~\cite{MartinBrualla2020GeLaTOGL} presented generative latent textured objects to synthesize images with a single 3D object.
Oechsle~\etal~\cite{Oechsle2019TextureFL} and Schwarz \etal~\cite{Schwarz2020GRAF} proposed to learn a mapping function from a 3D point to a color value, which is called a texture field and a radiance field, respectively, to synthesize images of a single 3D object.
Unfortunately, these methods focus on single 3D object and it is not trivial to extend them to handle 3D scenes with multiple objects of different classes.
Recently, Liao~\etal~\cite{Liao2020TULGM} introduced a generative model that generates multiple 3D objects and their textures.
However, their approach is limited to a small number of simple objects due to the complexity of the approach.

Material suggestion methods \cite{Chen2015Magicdecorator, Jain2012Materialmemex, Zhu2018ADA} aim to automatically assign texture maps to input 3D meshes by searching an external database.
However, they require a large-scale database of textured 3D models~\cite{Jain2012Materialmemex, Zhu2018ADA} or images with 3D material annotations~\cite{Chen2015Magicdecorator}, both of which are expensive to acquire.

\paragraph{Implicit representation}
Another relevant work to our method is implicit representation-based approaches.
These approaches use implicit representations or continuous representations, which are a class of learnable functions that map a coordinate to a certain type of signal, e.g., color and voxel occupancy.
Occupancy Networks~\cite{Mescheder2019OccupancyNL} and DeepSDF \cite{Park2019DeepSDFLC} reconstruct a 3D model by introducing an implicit function that takes a 3D coordinate as input and predicts the 3D occupancy of that position.
Oechsle~\etal~\cite{Oechsle2019TextureFL} learn a function that maps a 3D coordinate to color.
NeRF~\cite{Mildenhall2020NeRFRS} takes a 3D coordinate and a viewing direction as input and predict the images of an unobserved view point. 
Schwarz~\etal~\cite{Schwarz2020GRAF} designed a generative model based on NeRF.
Anokhin~\etal~\cite{Anokhin2020ImageGW} propose an image generator that independently calculates the color value at each pixel given a random vector and a 2D coordinate of that pixel.
Sitzmann~\etal~\cite{Sitzmann2020SIREN} showed that periodic activation functions can improve the representational performance.
Oechsle~\etal~\cite{Oechsle2019TextureFL} learn a function that maps 3D coordinate to color.
Peng~\etal~\cite{Peng2020ConvolutionalON} reconstruct a whole scene using a convolutional neural network that predicts occupancy.

For the implicit mapping between a coordinate to a desired output, encoding the coordinate is known to yield successful results. 
NeRF~\cite{Mildenhall2020NeRFRS} found that the sinusoidal positional encoding improves the representation power. Other approaches~\cite{Anokhin2020ImageGW, Tancik2020FourierFL} show that mapping Fourier features~\cite{Rahimi2007RandomFF} enables to learn high frequency functions.

Our approach is greatly inspired by these advances.
To effectively learn consistent color information of an input 3D scene from independently generated reference images, our method is designed with an implicit function that maps a 3D coordinate to an RGB color.
To enhance the image quality, we adopt the positional encoding.
In contrast to prior work, our method allows scene-level image synthesis and object manipulation.

%% file: section/method.tex
\section{Method}

\begin{figure*}[h!t]
\includegraphics[width=1.0\textwidth]{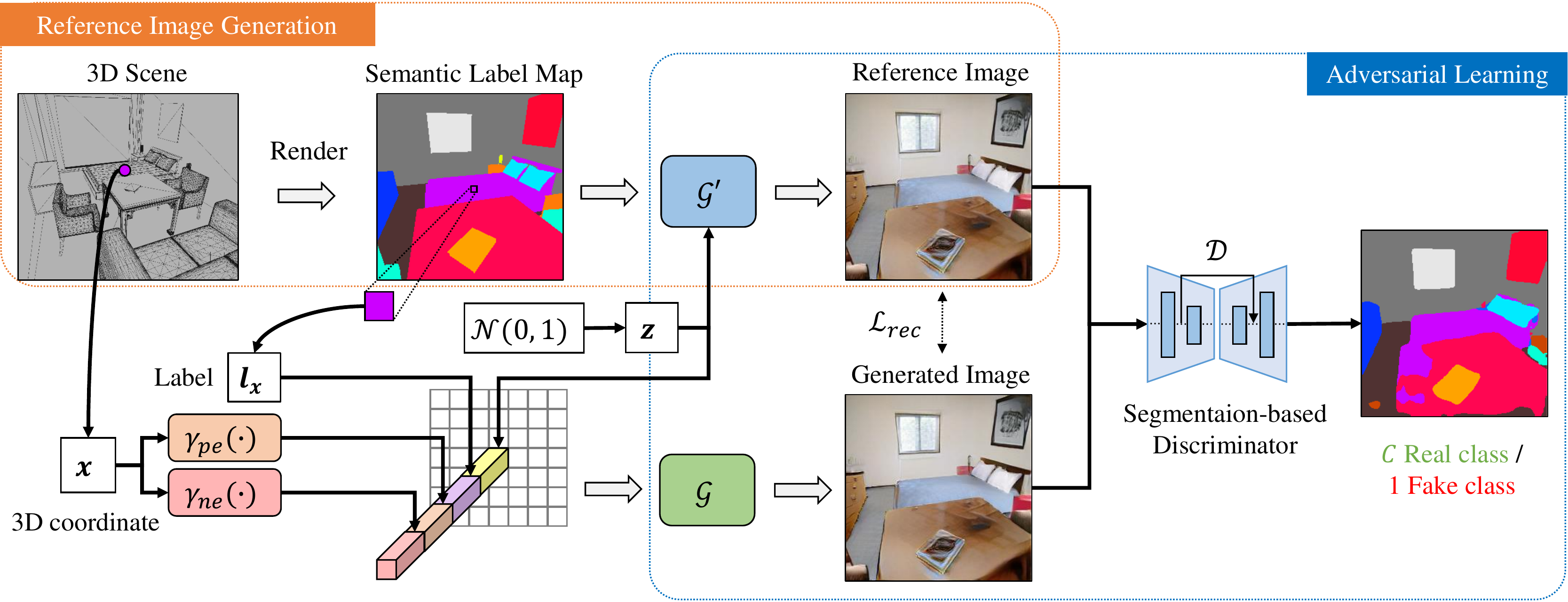}
\caption{ Overview of the proposed method. Our scene painting network $\mathcal{G}$ is trained with the reference images that are generated by per-view semantic image generator $\mathcal{G}'$.
$\mathcal{G}$ produces color from 3D coordinate $\boldsymbol{x}$, semantic label $\boldsymbol{l}_{\boldsymbol{x}}$ and style vector $\boldsymbol{z}$. The generated image is compared with the reference image. The segmentation-based discriminator classifies the authenticity of the generated images. Note that the losses are defined for \emph{each 3D point}, and multiple reference images generated by \emph{different viewpoints and style vector} teach $\mathcal{G}$ to paint the whole 3D points.
}
\label{fig:method}
\end{figure*}

In this section, we first provide an overview of our approach.
We then present detailed descriptions of the 3D scene painting network and its training.
\Fig{method} shows an overview of our 3D scene painting framework.
Our approach takes a 3D scene provided by a user as input.
An input 3D scene consists of 3D models with their semantic class labels.

\subsection{Reference Image Generator}

For training the painting network, we exploit an off-the-shelf semantic image synthesis network such as \cite{Park2019SemanticIS, Sushko2020YouON} that synthesizes an image from an input semantic segmentation map.
Specifically, during the training of the painting network, we sample multiple viewpoints and render 2D semantic segmentation maps of the input 3D scene for the sampled viewpoints.
We then feed the segmentation maps to the semantic image synthesis method to generate realistic-looking images.
We refer to the images as reference images as they will be used as reference labels for training our 3D scene painting network.
We also refer to the semantic image synthesis network as the reference image generator $\mathcal{G'}$.

The reference images from different viewpoints show inconsistent colors as they are generated independently.
Using the generated reference images, we train our painting network $\mathcal{G}$ to synthesize consistent color values.
To this end, we model $\mathcal{G}$ as a mapping from a 3D coordinate on a surface in the input 3D scene to a color value.
During the training of $\mathcal{G}$, we project 3D world coordinates in the input 3D scene using the viewpoints sampled for the reference images, and generate projected maps of 3D world coordinates.
Then, we feed the projected maps to $\mathcal{G}$ to generate images.
We train $\mathcal{G}$ to generate images similar to the reference images using a reconstruction loss $\mathcal{L}_{rec}$.
We also use an adversarial loss for the training of $\mathcal{G}$ so that $\mathcal{G}$ can reproduce high-frequency details in the reference images.

Our 3D scene painting network $\mathcal{G}$ supports different styles.
To achieve this, we use a semantic image synthesis method that takes a style vector as well as a semantic segmentation map as its input.
$\mathcal{G}$ also takes a style vector as input so that $\mathcal{G}$ can produce images of different styles with respect to the style vector.
Besides, motivated by OASIS~\cite{Sushko2020YouON}, $\mathcal{G}$ is also fed with semantic class labels to improve the image synthesis quality.

\subsection{3D Scene Painting Network}

Unlike conventional implicit representation-based approaches that learn a mapping function from a single 3D coordinate to a color value,
our 3D scene painting network $\mathcal{G}$ is designed to take a map of 3D coordinates and predict the colors of the coordinates.
This makes it possible for the network to incorporate information from the neighborhood of a pixel for the synthesis of an image, which enables more natural-looking synthesis results with view-dependent appearance changes.
In addition to 3D coordinates, $\mathcal{G}$ also takes semantic labels of the pixels and a style vector as input to enhance the synthesis quality based on class conditions and to support multiple styles.

Given a 3D scene consisting of multiple objects with semantic labels and a viewpoint, our image synthesis process using the scene painting network is as follows.
We first assemble 2D maps of the 3D world coordinates and the semantic labels by projecting the 3D points in the input scene for the given viewpoint.
Then, for each pixel, we compute the positional encodings of the projected 3D coordinate and concatenate them with the semantic label and the style vector to generate an input tensor $\mathbf{X}$.
Finally, we feed $\mathbf{X}$ to $\mathcal{G}$ to synthesize a realistic-looking image of the given viewpoint.

In detail, $\mathcal{G}$ is modeled as $\mathbf{I}=\mathcal{G}(\mathbf{X})$, where $\mathbf{I}$ is a resulting RGB image.
$\mathbf{X}~\in~\mathbb{R}^{H\times W\times D}$ is produced by concatenation of two types of positional encoding, $\gamma_{pe}$ and $\gamma_{ne}$, semantic label map $\boldsymbol{l}$, and a style vector $\boldsymbol{z}$ shared by all the pixels, i.e., $\mathbf{X}(\boldsymbol{p}) = [\gamma_{pe}(\boldsymbol{x}), \gamma_{ne}(\boldsymbol{x}), \boldsymbol{l}_{\boldsymbol{x}}, \boldsymbol{z}]$ where image coordinate $\boldsymbol{p}$ is a projection of 3D coordinate $\boldsymbol{x}=(x,y,z)$.
$D$ is sum of number of class $C$ and dimensions of $\gamma_{pe}$, $\gamma_{ne}$, $\boldsymbol{z}$. 
We model $\mathcal{G}$ as a convolutional neural network (CNN).
We refer the readers to our supplementary material for the detailed architecture of $\mathcal{G}$.

\paragraph{Positional encoding}
Our painting network $\mathcal{G}$ uses two functions $\gamma_{pe}$ and $\gamma_{ne}$ for encoding coordinates to enhance the image quality.
The first embedding function $\gamma_{pe}$ is the sinusoidal positional encoding function \cite{Mildenhall2020NeRFRS}.
For a 3D world coordinate $\boldsymbol{x}$, which is normalized into $[-1,1]$,
$\gamma_{pe}$ produces a $(6F+3)$-dimensional encoding vector where $F$ is a hyperparameter.
Specifically, we define $\gamma_{pe}$ as:
\begin{equation}
    \gamma_{pe}(\boldsymbol{x}) = [\boldsymbol{x}^\intercal, \gamma_{pe,0}(\boldsymbol{x}), \cdots, \gamma_{pe,F-1}(\boldsymbol{x})]^\intercal,
\label{eq:positional_encoding}
\end{equation}
where
\begin{equation}
\begin{split}
\gamma_{pe,f}(\boldsymbol{x})=[&\mathrm{sin}(2^f\pi x), \mathrm{cos}(2^f\pi x), \mathrm{sin}(2^f\pi y), \\
&\mathrm{cos}(2^f\pi y), \mathrm{sin}(2^f\pi z), \mathrm{cos}(2^f\pi z)].
\label{eq:positional_encoding_per_element}
\end{split}
\end{equation}

In practice, we observed that using only sinusoidal encoding is not enough to get high quality synthesis, as described in ablation studies. Therefore, we added additional nonlinear embedding $\gamma_{ne}$ that consists of two layers of MLP. The nonlinear positional embedding $\gamma_{ne}$ is defined as:
\begin{equation}
\gamma_{ne}(\boldsymbol{x}) = \boldsymbol{W_2}^{\intercal}[\sigma(\boldsymbol{W_1}^{\intercal}\boldsymbol{\hat{x}})^\intercal, 1]^\intercal, 
\label{eq:neural_embedding}
\end{equation}
where $\gamma_{ne}$ is a embedding function that is applied to $\boldsymbol{\hat{x}}$, $\boldsymbol{\hat{x}}$ is homogeneous coordinate of $\boldsymbol{x}$, and $\sigma$ is a nonlinear activation function.

\paragraph{Training}
For a given 3D scene, we train the painting network $\mathcal{G}$ using a reference image generator $\mathcal{G'}$.
Before the training, we create a pre-defined set of style vectors from normal distribution $\mathcal{N}(0,1)$ and a pre-defined set of viewpoints of the 3D scene.
Additionally, we create a semantic label map by projecting the semantic labels of the input 3D scene. From the semantic label map and the style vector $\boldsymbol{z}$, we synthesize a reference image $\boldsymbol{I}'$ using $\mathcal{G'}$.
At each iteration during the training, we sample $B$ pairs of viewpoints and style vector from the respective sets. 
In our experiments, we use a pre-trained generator network of OASIS~\cite{Sushko2020YouON} for $\mathcal{G'}$.

With the reference images, we train the painting network $\mathcal{G}$ by optimizing a reconstruction loss $\mathcal{L}_{rec}$ and an adversarial loss $\mathcal{L}_{adv}$.
The reconstruction loss $\mathcal{L}_{rec}$ is used for training $\mathcal{G}$ to synthesize images that resemble the reference images.
Specifically, $\mathcal{L}_{rec}$ is defined as:
\begin{equation}
    \mathcal{L}_{rec} = \frac{1}{K}\sum_{b=1}^{B} \left\| \mathcal{G}(\boldsymbol{X}_b) - \boldsymbol{I}'_b \right\|_1,
    \label{eq:L_G_recon}
\end{equation} 
where $K=BHW$ is a normalization factor. $H$ and $W$ are the height and width of a synthesized image.
$\boldsymbol{I'}_b$ is the $b$-th reference image in the current minibatch and $\boldsymbol{X}_b$ is the $b$-th input map.
$\mathcal{G}(\boldsymbol{X}_b)$ is an image generated using the same viewpoint and style vector as $\boldsymbol{I}'_b$. 

Because $\boldsymbol{I}'_b$s are generated independently, they are not consistent over the different views. 
Nonetheless, by minimizing Eq.~\eqref{eq:L_G_recon} for multiple reference images, we can aggregate such inconsistent images and train $\mathcal{G}$ to synthesize consistent images. 

As minimizing the reconstruction loss $\mathcal{L}_{rec}$ aggregates information from inconsistent reference images, it makes $\mathcal{G}$ to produce blurry images.
To alleviate this, we adopt the adversarial learning to guide $\mathcal{G}$ to synthesize more realistic-looking images with high-frequency details.
Specifically, we adopt the semantic segmentation-based adversarial learning approach of OASIS~\cite{Sushko2020YouON}
where a discriminator learns to classify each pixel in an input image into $(C+1)$ semantic classes including one fake class. In our framework, the ground truth of label map for the discriminator is semantic label map of the input 3D scene.
The adversarial loss $\mathcal{L}_{adv}$ is defined as:
\begin{equation}
\mathcal{L}_{adv} = -\frac{1}{K} \sum_{b=1}^{B}
    \sum_{c=1}^{C}\alpha_{c} 
        \sum_{i=1}^{H\times W} \boldsymbol{L}_{b,i,c}
            \log{\mathcal{D}\left(
                \boldsymbol{I}_{b}\right)_{i,c}}.
\label{eq:L_G_adv}
\end{equation}
where $\alpha_c$ is a weight for each class to resolve the class imbalance problem, which is borrowed from ~\cite{Sushko2020YouON}.
$\mathcal{D}$ is a semantic segmentation-based discriminator.
$\boldsymbol{L}_{b,i,c}$ is the $c$-th element of the semantic class label represented as a one-hot vector at the $i$-th pixel of the $b$-th image, which is either $0$ or $1$.
Similarly, $\mathcal{D}(\boldsymbol{I}_b)_{i,c}$ is the $c$-th element at the $i$-th pixel of the discriminator output $\mathcal{D}(\boldsymbol{I}_b)$.

The discriminator $\mathcal{D}$ is trained using a loss defined as:
\begin{equation}
\begin{split}
\mathcal{L}_{\mathcal{D}} = &-\frac{1}{K} \sum_{b=1}^{B} 
\sum_{c=1}^{C}\alpha_{c}      
    \sum_{i=1}^{H\times W} \boldsymbol{L}_{b,i,c}
        \log{\mathcal{D}\left(
            \boldsymbol{I}'_b \right)_{i,c}} \\
     &-\frac{1}{K} \sum_{b}^{B} 
    \sum_{i=1}^{H\times W} 
        \log{ \mathcal{D}\left(
            {\boldsymbol{I}_b}\right)_{i,C+1}}.
    \label{eq:L_D_adv}
\end{split}
\end{equation}
where $C+1$ indicates the fake class label.
In our experiments, we initialize $\mathcal{D}$ with a pre-trained discriminator from OASIS~\cite{Sushko2020YouON} to accelerate the learning. 
For effective adversarial learning using semantic class labels, we also adopt the LabelMix regularizer of OASIS~\cite{Sushko2020YouON}.
We refer the readers to \cite{Sushko2020YouON} for more details.

Our final loss for training a 3D scene painting network $\mathcal{G}$ is then defined as:
\begin{equation}
    \mathcal{L}_{\mathcal{G}} = \mathcal{L}_{recon} + \lambda_{adv}\mathcal{L}_{adv}
    \label{eq:L_G}
\end{equation}
where $\lambda_{adv}$ is the weight for the adversarial loss.

%% file: section/experiment.tex
\section{Experiments}

\subsection{Implementation details}
\noindent\textbf{Network.}
We implemented the generator as a CNN. It consists of three convolutional layers with $512$ feature dimensions. The convolution kernel size is $3$, padding is $1$, and there is no stride. The activation function is leaky-ReLU~\cite{Maas2013RectifierNI} and its negative slope is $0.2$.
The hyperbolic tangent function is applied to the last output. The dimensions of positional augmentation is set to $Dim(\gamma_{pe})=15$ and $Dim(\gamma_{ne})=64$. The dimension of $z$ is $64$. The number of classes $C$ is 150, which is equivalent to ADE20K~\cite{Zhou2017ADE20K} dataset.
For the training, we used Adam~\cite{Kingma2015AdamAM} optimizer, the learning rate of generator is $1e^{-3}$, the learning rate of discriminator is $1e^{-4}$. We set $\lambda_{0}=10$ and $\lambda_{1}=10$ in our experiment. Unless otherwise mentioned, we use the aforementioned settings in the experiments shown in this paper.

\vspace{2mm}\noindent\textbf{Data preparation.}
For the experiment, we utilize 3D models and their arrangements from SceneNet~\cite{Handa2016SceneNet} dataset. We use Blender~\cite{Blender} to render semantic label maps and depth maps from various viewpoints. With depth maps and intrinsic parameter of the synthetic camera, we acquire 3D coordinates of every pixels. With this procedure, user can generate semantic label maps, coordinate images and reference images by changing view points. We used bedroom, kitchen, living room and office scenes. For each scene, there are 98, 62, 94, 137 objects and we used 500, 400, 500, 600 viewpoints in the scene. 
We generate reference images from semantic label maps using pre-trained OASIS~\cite{Sushko2020YouON} model. As the dataset has no label on the objects, we assign each 3D model to be one class of $150$ classes as in ADE20k~\cite{Zhou2017ADE20K} dataset and OASIS~\cite{Sushko2020YouON}\footnote{Our annotation will be published once accepted.}. 

\vspace{2mm}\noindent\textbf{Toy example scene}
For quantitative experiments including ablation studies, we created a toy example scene by modifying a bedroom scene of SceneNet~\cite{Handa2016SceneNet}. The scene is of a room containing 14 classes. In ablations, we used 9 different style vectors to support 9 scene styles. We sampled the 9 vectors and fixed them before training.

\begin{figure*}[t]
\includegraphics[width=1.0\textwidth]{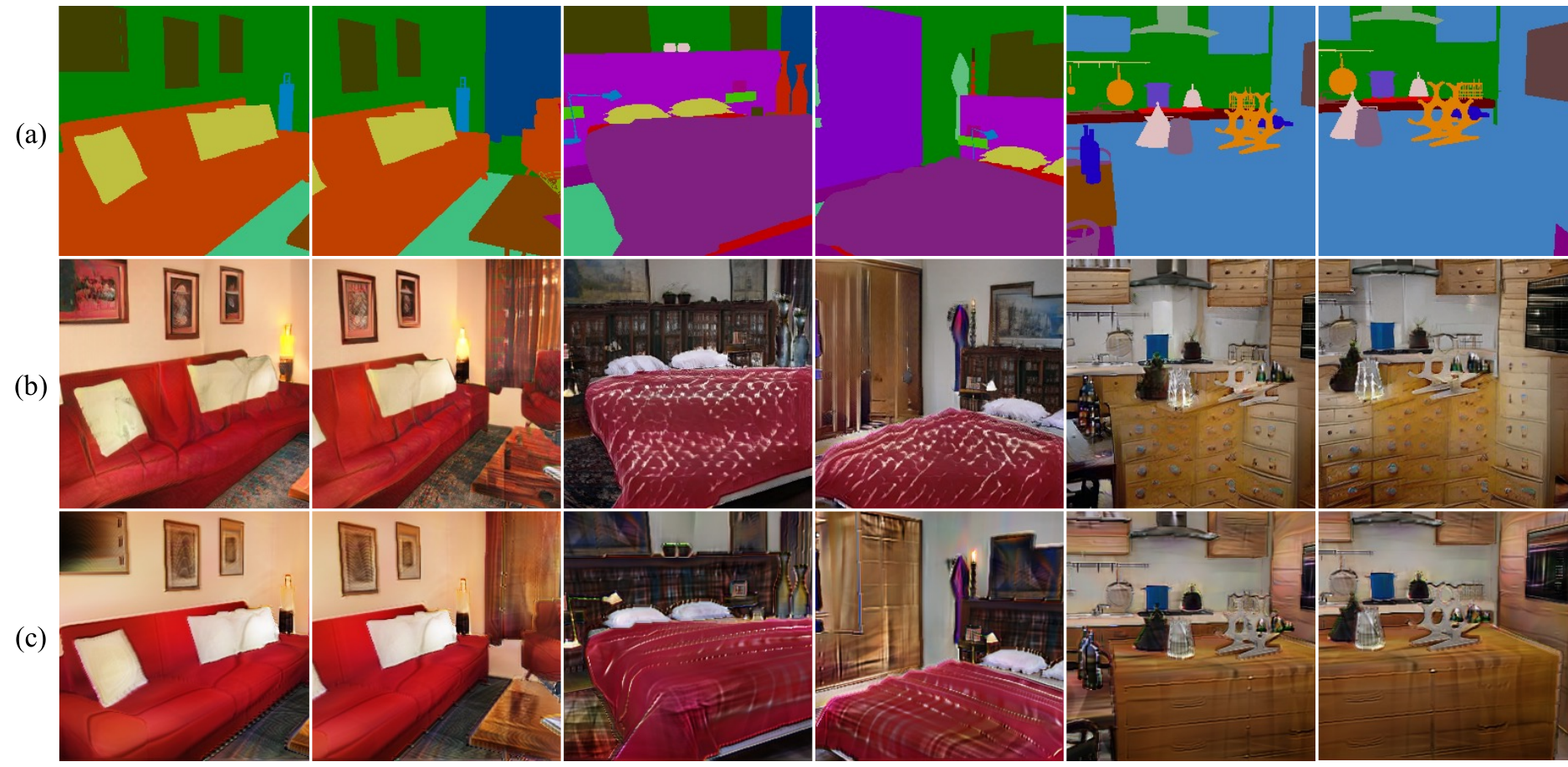}
\caption{We compared images generated by our method and the images produced by OASIS~\cite{Sushko2020YouON} on the three scenes. For each scene, two images from two different views were compared. (a) Semantic label map, (b) image produced by OASIS, and (c) our results. Our method can generate more coherent images. 
}
\vspace{2mm}
\label{fig:result_oasis_ours}
\end{figure*}

\subsection{Evaluation metrics}

We evaluate the quality of our results using the following metrics.

\vspace{2mm}
\noindent\textbf{Frechet Inception Distance (FID).} FID~\cite{Heusel2017FID}
represents distance between two distribution of images.
We measure FID between the generated images and real-images from ADE20k dataset~\cite{Zhou2017ADE20K}. The small value indicates a generated image is close to the real-image.

\vspace{2mm}
\noindent\textbf{Mean Intersection over Union (mIoU).} To measure how similar the generated image to real images, we computed mIoU between the prediction of generated image with pre-trained semantic segmentation network~\cite{Xiao2018UPerNet} and semantic label map that was used to generate the images. mIoU score is high when the generated images are realistic.

\vspace{2mm}
\noindent\textbf{View Consistency (VC).} VC measures how our images are coherent over the different views. A common choice of measuring view consistency of images is a reference metric such as PSNR. However, we cannot measure reference metric as ground truth multi-view images are unavailable in our setting.
Since we know 3D coordinates of every pixel in the generated images, we can compare pixel color of points in local neighborhood. 
We define local neighborhood $N_{\boldsymbol{p}}$ as a cell in 3D grid whose center is point $\boldsymbol{p}$ and $\mathcal{C}(N_{\boldsymbol{p}})$ as set of pixel colors correspond to $N_{\boldsymbol{p}}$. \footnote{We round coordinates to obtain local neighborhood.} 
We define the View Consistency (VC) metric as follows:
\begin{equation} 
\frac{1}{n(\mathcal{N})} \sum_{N_{\boldsymbol{p}}\in{}\mathcal{N}} \max_{\boldsymbol{y}\in{}\mathcal{C}(N_{\boldsymbol{p}}), \boldsymbol{y}'\in{}\mathcal{C}(N_{\boldsymbol{p}})} | \boldsymbol{y} - \boldsymbol{y'} |_{2}
\label{eq:measure_consistency}
\end{equation}
where $(\boldsymbol{y},\boldsymbol{y}')$ are a pair of color values from the generated images. 
The pair corresponds to the same local neighborhood $N_{\boldsymbol{p}}$.
$\mathcal{N}=\{N_{\boldsymbol{p}}|n(\mathcal{C}(N_{\boldsymbol{p}}))>=2\}$ is the set of the local neighborhoods having two or more corresponding pixels.
Low VC measure indicate high view consistency of the measured images.

\begin{figure*}[t]
\includegraphics[width=1.0\textwidth]{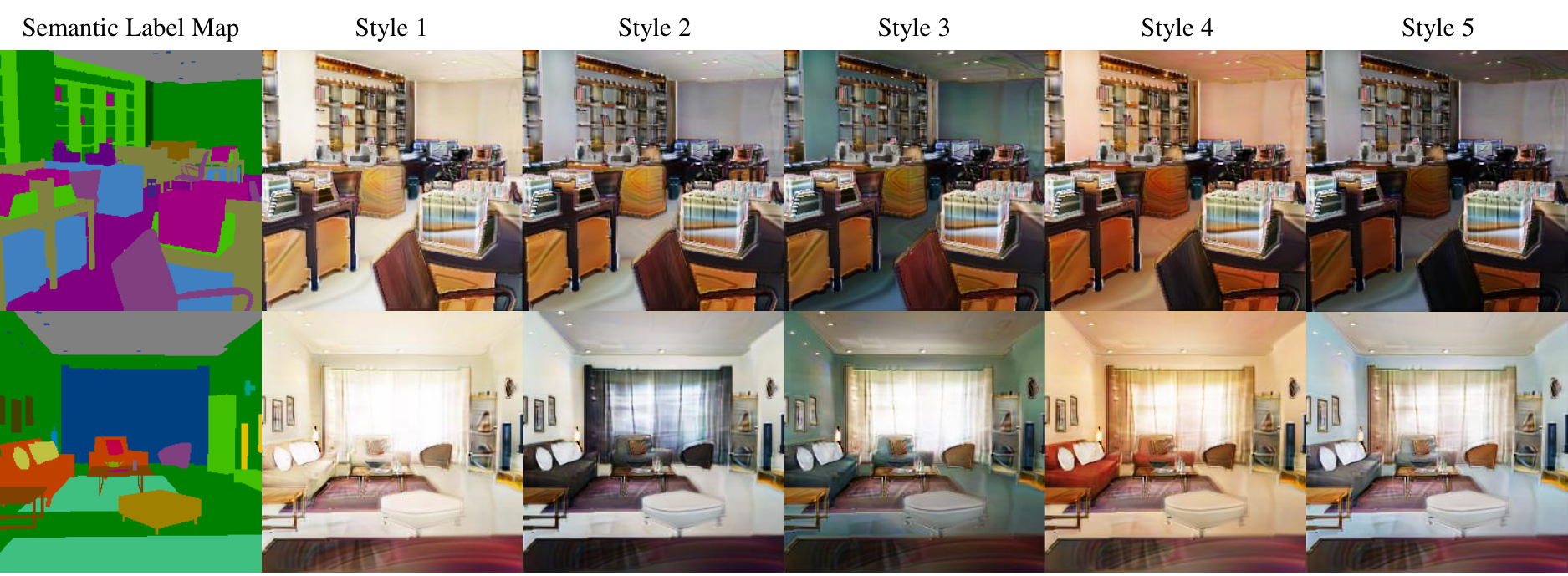}
\caption{Examples of style manipulation using our image generator.}
\vspace{2mm}
\label{fig:result_style}
\end{figure*}

\begin{figure*}[t]
\includegraphics[width=1.0\textwidth]{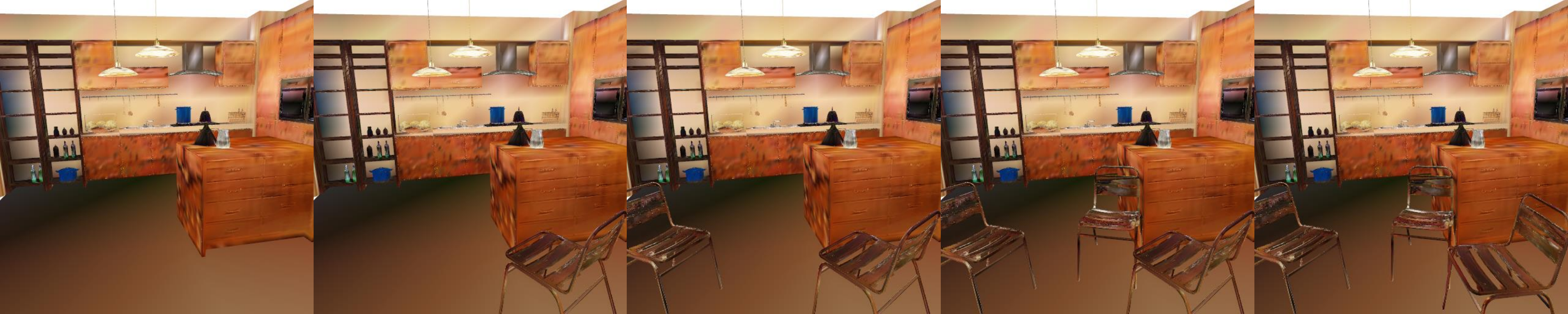}
\caption{
Colored meshes generated with our method and scene editing examples. For scenes in the first column to the fourth column, we progressively placed a chair into the scene, and in the last column we changed the pose of one chair in the fourth scene.
}
\label{fig:result_edit}
\end{figure*}

\subsection{Qualitative results}

\vspace{2mm} \noindent\textbf{Image quality and View consistency.}
We visually compare the quality and view consistency of our results with images generated by OASIS~\cite{Sushko2020YouON}. 
OASIS produces a geometrically invalid output, especially in the scene with complex geometry, which is difficult to infer from semantic maps as shown by the last two columns of \Fig{result_oasis_ours} (a) and (b). Moreover, the content is not view-consistent. 
In contrast, our method produces geometrically valid and view consistent as shown by \Fig{result_oasis_ours} (c).

\vspace{2mm} \noindent\textbf{Scene style control.}
Our approach readily controls the style of a scene by changing the style vector $\boldsymbol{z}$ without re-training.
For each style vector, we generated images of scenes. \Fig{result_style} shows examples of change of scene styles.

\vspace{2mm} \noindent\textbf{Mesh color generation and 3D scene editing.}
We show that our approach can be used to generate color of mesh which allows scene editing using 3D graphics tools.
We generate colored meshes by back-projecting the generated image into the 3D scene and assigned vertex color of the 3d meshes.
\Fig{result_edit} shows example images rendered from the edited scene with a colored mesh generated by our method.
This feature also allows users to modify scene color and configuration more easily.

\subsection{Quantitative Results}
To the best of our knowledge, our approach is the first attempt at painting 3D scenes from a configurable 3D scene layout. Therefore, we conduct experiments on quantitative performance to evaluate the effectiveness of the proposed method. We conduct ablation studies by using the toy example scene. Additionally, we compare our model with OASIS~\cite{Sushko2020YouON} which is our reference image generator.

\vspace{2mm}
\noindent\textbf{Architecture selection of generator.}
We conducted experiments on two possible architectures, such as MLP and CNN as the image generator. We make both architectures have a similar number of parameters (MLP: $1.9$M, CNN: $1.7$M\footnote{The number of layers of CNN is $5$ and hidden dimension is $192$ while the number of layers of MLP is $7$ and hidden dimension is $512$.}) for a fair comparison. Table~\ref{tab:generator} shows the results. The results show a trade-off between them in terms of view-consistency and single image quality. For view-consistency, MLP achieves a lower consistency loss than CNN. For single image quality, CNN achieves the better mIoU, FID, and VC scores. This phenomenon happens because MLP directly maps coordinate information to the RGB value, whereas CNN has a weaker consistency by minimizing Eq.~(\ref{eq:L_G_recon}).

\begin{table}[h]
\caption{Comparison on generator architectures.}
\centering
\resizebox{0.75\linewidth}{!}{
\begin{tabular}{c|ccc}
\toprule
\multirow{2}{*}{Architecture} & \multicolumn{3}{c}{Measure} \\
\cline{2-4}
 & mIoU~($\uparrow$) & FID~($\downarrow$) & VC~($\downarrow$) \\ 
\midrule
MLP & 0.488 & 147.989 & \textbf{25.447} \\
CNN & \textbf{0.557} & \textbf{113.419} & 39.069 \\
\bottomrule
\end{tabular}%
}%
\label{tab:generator}%
\end{table}%

\begin{figure*}[t]
\includegraphics[width=1.0\textwidth]{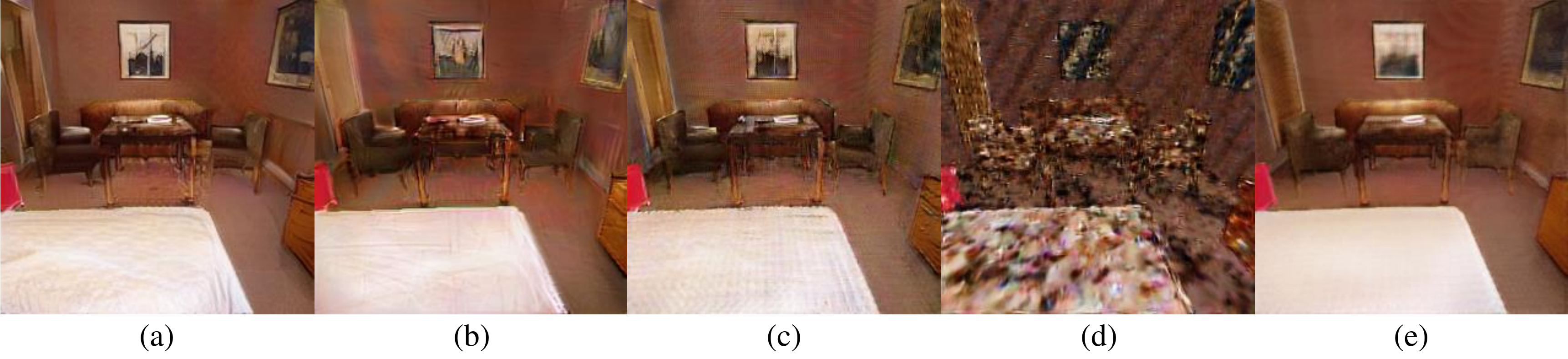}
\caption{The examples of positional encoding artifacts. (a) A reference image generated from OASIS~\cite{Sushko2020YouON} (b) The result of $\gamma_{pe}(15),\gamma_{ne}(64)$. (c) The result of $\gamma_{pe}(63),\gamma_{ne}(64)$. (d) The result of Fourier$(64)$ encoding. (e) The result of the network that train with only the reconstruction loss. Images of (b), (c), (d) are generated by the networks trained with the full loss.} 
\label{fig:result_artifact}
\end{figure*}

\vspace{2mm} \noindent\textbf{Positional encoding.}
We observe that the proper positional encoding improve the quality of the generated images. In Table~\ref{tab:pos_enc}, we experimented with three different positional encoding schemes: sinusoidal~\cite{Mildenhall2020NeRFRS} encoding $\gamma_{pe}$ in Eq.~(\ref{eq:positional_encoding}), Fourier~\cite{Tancik2020FourierFL}\footnote{
It is defined as: 
$ \gamma(\boldsymbol{x}) = [\mathrm{sin}(\boldsymbol{b}^{\intercal}\boldsymbol{x}),  \mathrm{cos}(\boldsymbol{b}^{\intercal}\boldsymbol{x})]^{\intercal}$,
where $\boldsymbol{b}$ is random vector sampled from normal distribution $\mathcal{N}(0,1)$.}, and nonlinear embedding $\gamma_{ne}$ in Eq.~(\ref{eq:neural_embedding}). The value in parenthesis in the encoding function are the output dimension of the function.
The baseline is the generator with no positional encoding (without using $\gamma_{pe}$ and $\gamma_{ne}$).
Finally, we found that using $(\gamma_{pe}(15),\gamma_{ne}(64))$ shows lower FID than that of Fourier or the baseline.
In detail, using only sinusoidal or Fourier encoding induces an artifact in the resulting image. As shown in \Fig{result_artifact}~(c), sinusoidal encoding with a high number of dimensions shows a grid-like artifact. The number of dimension of $\gamma_{pe}$ significantly affect the result. When the number of dimensions is high, the frequency of features gets high. Such phenomenon makes each position cannot be clearly distinguished. When we reduce the number of dimensions of $\gamma_{pe}$, as shown in \Fig{result_artifact}~(b), the generated images do not show the artifact. When we compensate the low dimension with $\gamma_{ne}$ that has proper output dimension, the mIoU and FID score are improved. 
\Fig{result_artifact}~(d) results from Fourier encoding, and it shows sinusoidal artifact. 

\begin{table}[h]
\caption{Evaluation on various positional encoding approaches. $\gamma_{pe}(dim)$ or $\gamma_{ne}(dim)$ indicate dimension of each vectors.}
\centering
\resizebox{0.75\linewidth}{!}{
\begin{tabular}{c|cccc}
\toprule
\multirow{2}{*}{Positional encoding} & \multicolumn{2}{c}{Measure} \\
\cline{2-3}
 & mIoU~($\uparrow$) & FID~($\downarrow$) \\ 
\midrule
None & 0.498 & 129.989 \\
Fourier$(64)$ & 0.217 & 343.43 \\
$\gamma_{ne}(128)$ & 0.533 & 126.086 \\
$\gamma_{pe}(63)$ & 0.523 & 130.147 \\
$\gamma_{pe}(63),\gamma_{ne}(64)$ & 0.526 & 128.894 \\
$\gamma_{pe}(39),\gamma_{ne}(64)$ & 0.55 & 130.633 \\
$\gamma_{pe}(15),\gamma_{ne}(64)$ & \textbf{0.557} & \textbf{113.419} \\
$\gamma_{pe}(15),\gamma_{ne}(128)$ & 0.509 & 118.714 \\
$\gamma_{pe}(15)$ & 0.514 & 119.85 \\
\bottomrule
\end{tabular}
}
\label{tab:pos_enc}
\end{table}%

\vspace{2mm} \noindent\textbf{Loss function.}
In this experiment, we showed the necessity of adversarial loss.
We experimented with the generator by using different configurations of loss functions: reconstruction loss (Eq.~\eqref{eq:L_G_recon}), adversarial loss (Eq.~\eqref{eq:L_G_adv}, Eq.~\eqref{eq:L_D_adv}) and full loss (Eq.~\eqref{eq:L_D_adv}, Eq.~\eqref{eq:L_G}).

Table~\ref{tab:loss} shows the result.
The reconstruction loss improves perceptual quality and view consistency but showed lower mIoU. \Fig{result_artifact} (e) shows the result of reconstruction loss, and the image has a monotonic texture compared to the results of full loss, shown in \Fig{result_artifact} (b) and (c).
The adversarial loss improves perceptual quality, as indicated by improved mIoU. %
The results using full loss function shows improved perceptual quality, as demonstrated by mIoU and FID, and improved view consistency as shown by improved VC.

\begin{table}[h]
\caption{Evaluation of loss functions.}
\centering
\resizebox{0.85\linewidth}{!}{
\begin{tabular}{c|ccc}
\toprule
\multirow{2}{*}{Loss function} & \multicolumn{3}{c}{Measure} \\
\cline{2-4}
    & mIoU~($\uparrow$) & FID~($\downarrow$) & VC~($\downarrow$) \\
\midrule
Reconstruction only & 0.408 & 123.548 & 49.854 \\
Adversarial only & 0.462 & 154.15 & 59.563 \\
Full & \textbf{0.557} & \textbf{113.419} & \textbf{39.069} \\
\bottomrule
\end{tabular}
}
\label{tab:loss}
\end{table}%

\vspace{2mm} \noindent\textbf{Image quality and View consistency.}
We first respectfully emphasize that our goal significantly differs from that of the image synthesis methods, and it is not fair to compare our method against previous ones solely based on conventional metrics.
Nevertheless, we conducted an experiment with recent image synthesis approaches on the toy example scene.
In Table~\ref{tab:method}, among the image synthesis methods, OASIS~\cite{Sushko2020YouON} performs the best.
The performance of our method is bounded by the reference image generator, thus our results are inferior to those of OASIS in terms of mIoU and FID. 
Despite this, our method shows better mIoU and FID scores compared to the other recent approaches, and shows the best view consistency (VC).
OASIS's perceptual quality is better than ours, but OASIS is hard to make coherent images for the different viewpoints.
Additionally, we also compared different methods as our reference generator for the analysis of the upper bound of our method.
The table~\ref{tab:method} shows that Ours+OASIS shows better performance than Ours+SPADE.
This justifies our choice of the reference image generator.

\begin{table}[h]
\caption{Comparison to semantic image synthesis methods.}
\centering
\resizebox{0.995\linewidth}{!}{
\begin{tabular}{c|c|ccc}
\toprule
\multicolumn{2}{c}{Method} & \multicolumn{3}{c}{Measure} \\
\cline{1-5} 
Category & Name & mIoU~($\uparrow$) & FID~($\downarrow$) & VC~($\downarrow$) \\ 
\midrule
 & OASIS~\cite{Sushko2020YouON_arxiv} & \textbf{0.622} & \textbf{94.258} & \underline{62.694} \\
 & SPADE~\cite{Park2019SemanticIS} & 0.508 & 129.273 & 83.899 \\
Image synthesis & CC-FPSE~\cite{Liu2019LearningTP} & 0.544 & 150.408 & 73.797 \\
 & SESAME~\cite{Ntavelis2020SESAMESE} & 0.529 & 140.996 & 72.118 \\
 & LGGAN~\cite{Tang2020LocalCA} & 0.541 & 129.839 & 72.219 \\
\midrule
\multirow{2}{*}{\textbf{3D scene painting}} & Ours + SPADE & 0.419 & 156.999 & 52.347 \\
 & Ours + OASIS & \underline{0.557} & \underline{113.419} & \textbf{39.069} \\
\bottomrule
\end{tabular}%
}%
\label{tab:method}
\end{table}%

%% file: section/limitation.tex
\section{Limitation}
The proposed approach can produce coherent images, and it be combined with any image generation approach. Albeit such benefits, our results depend on the quality of the reference image generator. 
Besides, our approach hardly generates transparent objects such as a window or view-dependent appearances. This limitation stems from the input representation. As the network assigns one color for each 3D coordinate, it cannot model view-dependent lighting effect. For transparency, the model treats all objects as opaque because we only feed the front most coordinate to the network. 
Our approach takes about five hours for one toy example scene with four Titan Xp GPUs. For larger scenes such as SceneNet, we train the network for about 15 hours for one scene. Since our generator is scene-dependent, further work is required for reducing the training time.

%% file: section/conclusion.tex
\section{Conclusion}
We proposed an image synthesis method that synthesizes color of configurable 3D scene layout and a training scheme that does not require 3D color supervision. Given 3D coordinate and semantic label map, our scene painting network synthesizes realistic and view-consistent images. Our method ensures the view-consistency of synthesis which is not addressed in semantic image synthesis method.
In addition, our method can be used to generate color of a scene containing multiple objects which also allows users to modify scene color and configuration using 3D graphics tools.

%% file: section/supplement.tex
\begin{figure*}[b]
\centering
\includegraphics[width=\linewidth]{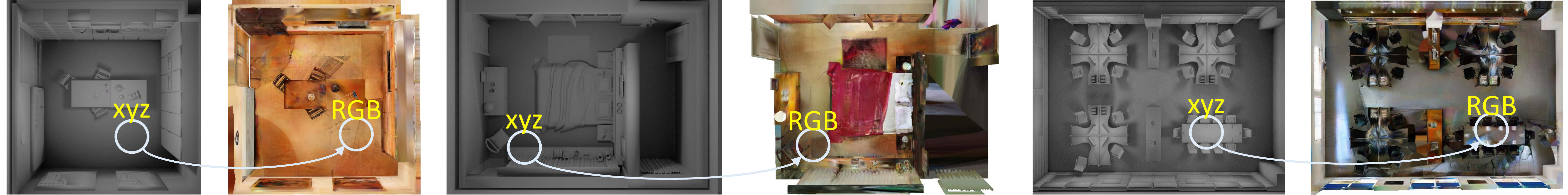}
\caption{Our approach learns to map 3D coordinates and semantic labels to RGB colors. For each of the three examples, the left and right images show an input 3D geometry and our scene coloring results shown in the main paper, respectively.} 
\label{fig:scene_coloring}
\end{figure*}

This supplementary material provides more qualitative results, details of the network architecture, analysis of positional encoding and some failure cases that could not be included in the main paper due to the limit of space. Especially, we provide a video that shows the visual consistency and the advantages of our scene painting framework.

Figure~\ref{fig:scene_coloring} shows the output of our work and explains that our method generates scene dependent colorization. The coloring process assigns color for every 3D point in the scene. Here, style vector $\boldsymbol{z}$ is fixed for a scene.  After the scene coloring, the scene can be edited based on the underlying geometry, e.g., the objects can be relocated.

\section{Qualitative Results}

\subsection{Video}
The supplementary video presents a comparison between OASIS~\cite{Sushko2020YouON} and our method. The results of OASIS in the video are generated by applying OASIS to semantic label maps in a frame-wise manner. While the results of OASIS show temporal inconsistency and distorted structures as it does not consider underlying 3D structures, our results show consistent and geometrically accurately synthesized frames. The video also presents scene editing examples using our method.

\subsection{Scene editing}

With recent semantic image synthesis methods such as OASIS~\cite{Sushko2020YouON}, it is challenging to accurately adjust image contents such as the positions and orientations of objects. However, as our scene painting network learns a mapping from a 3D coordinate to a color space, we can exploit the mapping to edit the layout of 3D objects in an input scene. Figure~\ref{fig:fig_scene_editing} shows scene editing examples. Video examples are also provided in the supplementary video.

\begin{figure}[h]
\centering
\includegraphics[width=\linewidth]{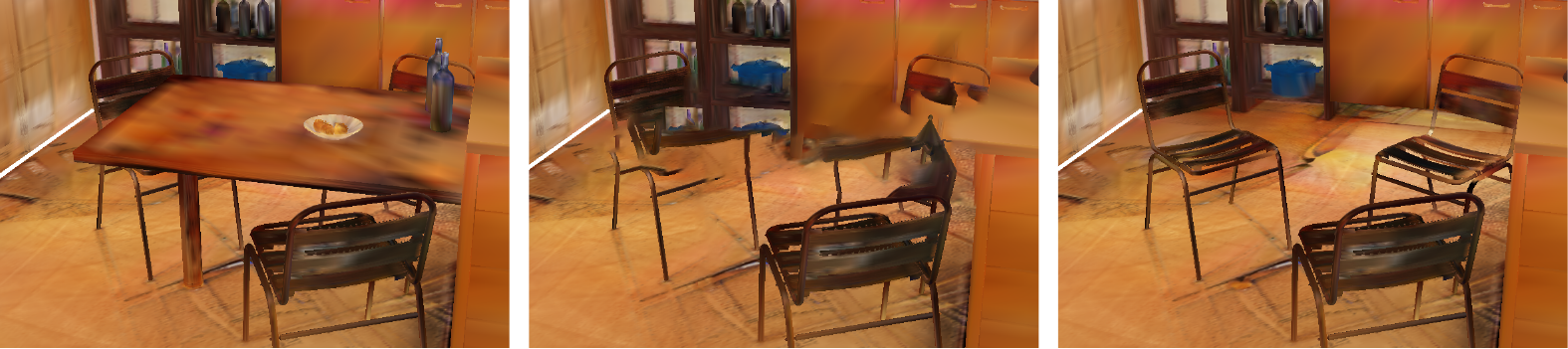}
\caption{
Comparison with an image editing method. (Left) colored scene, (Middle) Object removal using Adobe Photoshop Content-Aware Fill, (Right) our result.
}
\label{fig:image_editing}
\end{figure}

Please note that image editing methods can produce similar result to our scene editing results.
However, as our method does not edit an image directly it is not appropriate to compare our method with image editing methods.
Nonetheless, we provide comparison with an image editing method in Figure~\ref{fig:image_editing}.
Image editing methods cannot produce geometrically valid images, whereas our scene editing can produce view consistent and geometrically valid images.

\section{Network Architecture}
Table 1 in the main paper compares the performances of different network architectures for the scene painting network: a MLP-based generator and a CNN-based generator.
In this section, we present their detailed architectures.

\subsection{Architecture Details}

\begin{figure}[t!]
\centering
\includegraphics[width=1.0\linewidth]{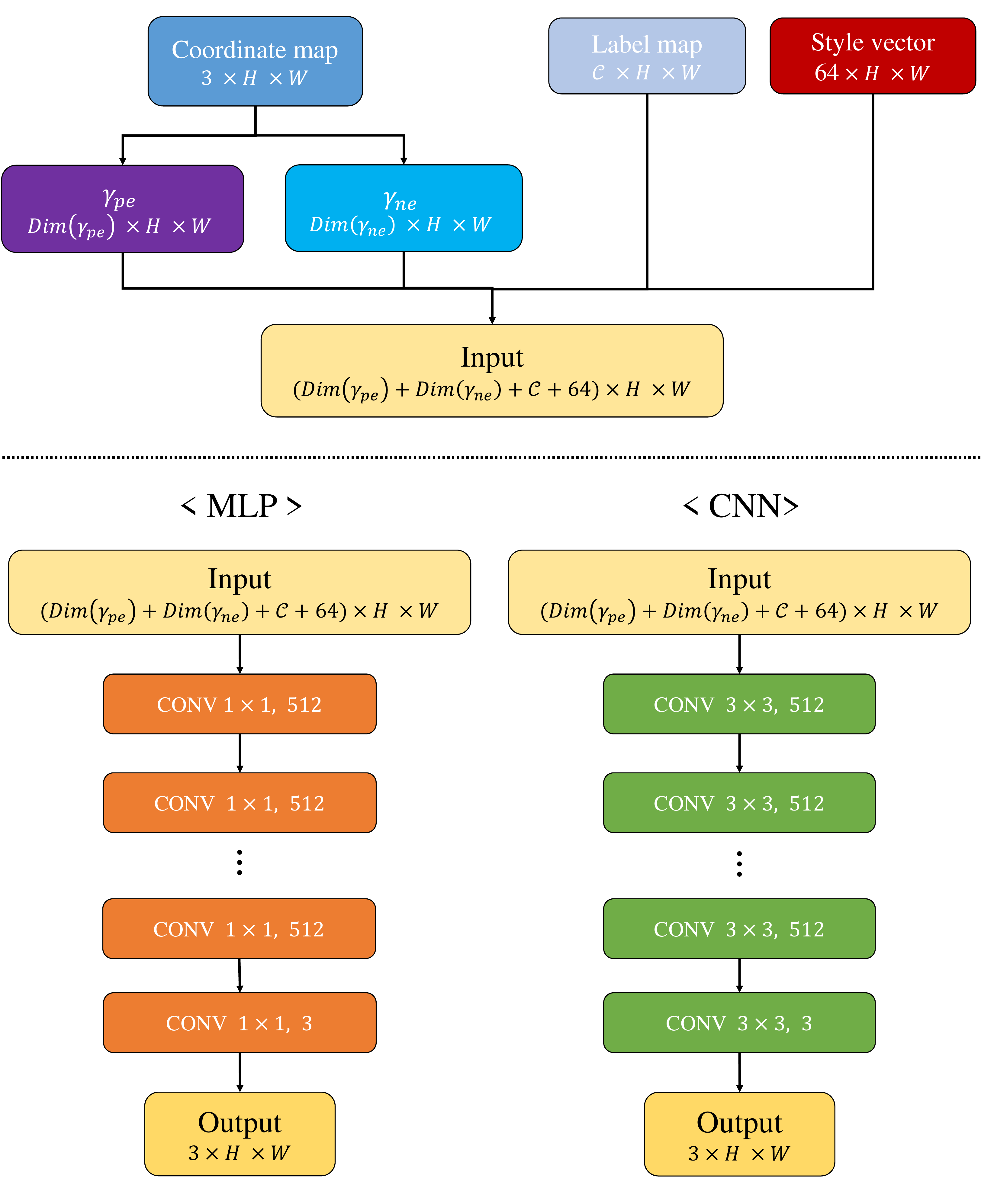}
\caption{Architectures of the MLP and CNN generators. `CONV $3\times3, 512$ means a convolution layer with kernel size $3\times 3$ and with 512 output channels.} 
\label{fig:arch_flowchart}
\end{figure}

\begin{figure*}[t!]
\centering
\includegraphics[width=0.9\linewidth]{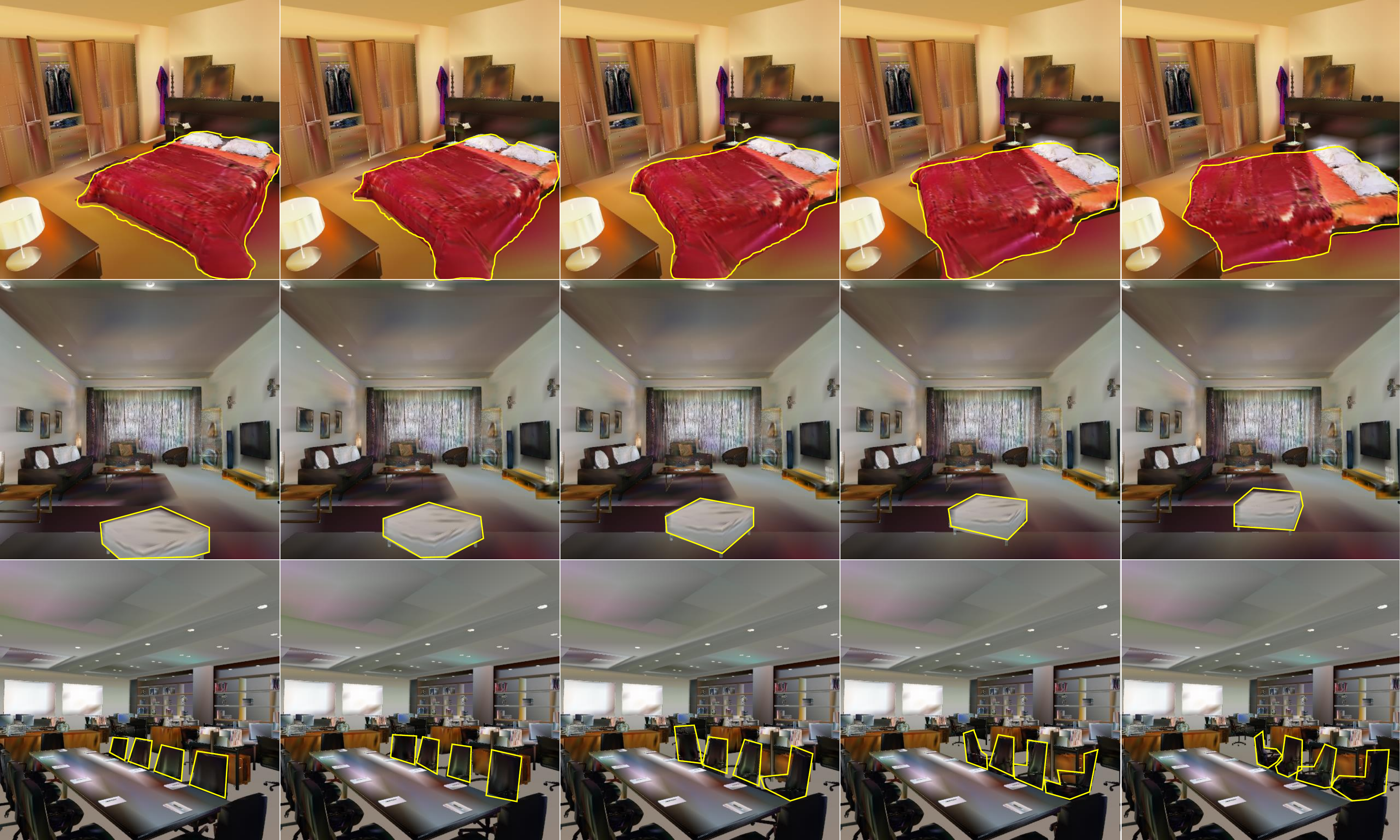}
\caption{Examples of scene editing. Edited objects are marked with yellow boundaries. Top: changing the position and orientation of the bed. Middle: changing the position of a stool. Bottom: changing the positions and orientations of chairs.} 
\label{fig:fig_scene_editing}
\end{figure*}

In Figure~\ref{fig:arch_flowchart}, we show the network architectures of the MLP and CNN-based generators. For both networks, we assume that a 3D scene with semantically labeled objects and a style vector is given as input.
From a given 3D scene with semantically labeled objects, we render a 2D semantic label map and a 3D world coordinate map, which is normalized into $[-1,1]$.
We compute the positional encodings from the coordinate map $\gamma_{pe}$ and $\gamma_{ne}$.
A style vector is a 64-dimensional vector sampled from the standard normal distribution.
We spatially stack the style vector and obtain a style map of the spatial size $H \times W$, which is the same as the label map's size.
Then, we concatenate the positional encodings, the label map, and the style map to generate an input tensor for the generator.
The MLP generator is implemented with $1 \times 1$ convolution layers as it is equivalent to applying fully connected layers to each pixel independently.
The CNN generator is implemented with $3 \times 3$ convolution layers. There is no down-sampling in both generators.
Every convolution layer except for the last one is followed by a leaky ReLU~\cite{Maas2013RectifierNI} layer.
After the last convolution layer, we have a hyperbolic tangent layer in both networks.
Both networks output a synthesized 2D image.
In our experiments, we set the dimensions of the positional encodings to $Dim(\gamma_{pe})=15$ and $Dim(\gamma_{ne})=64$.
The dimension of the style vector is 64.
The number of classes $C$ is set to 150, which is equivalent to the number of classes in the ADE20K~\cite{Zhou2017ADE20K} dataset.

\begin{figure*}[t!]
\includegraphics[width=1.0\linewidth]{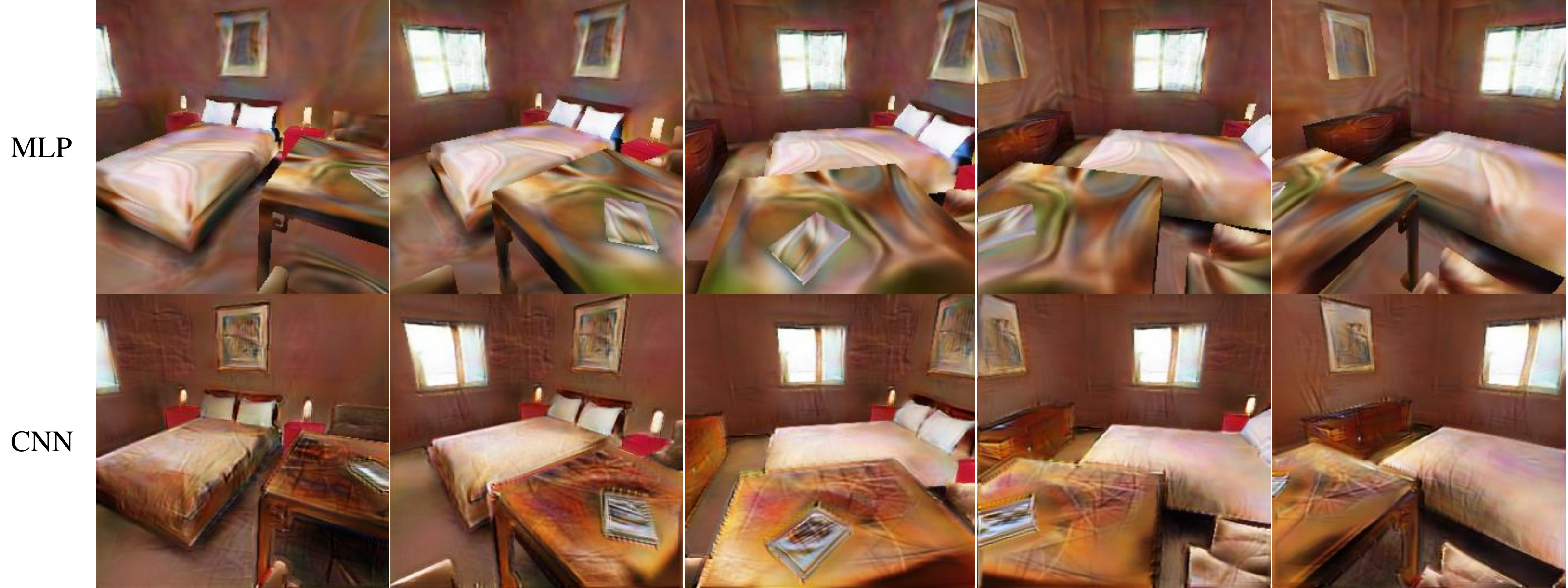}
\caption{Comparison of the MLP and CNN generators. The top and bottom rows show generated images from the MLP and CNN generators, respectively.} 
\label{fig:mlp_cnn}

\end{figure*}

\subsection{MLP Generator vs.~CNN Generator}
\Fig{mlp_cnn} shows a qualitative comparison between the MLP and CNN generators.
As shown in the figure, the MLP generator achieves better consistency between different viewpoints than the CNN generator.
On the other hand, the CNN generator synthesizes more natural-looking images with fewer artifacts and more natural-looking shading. %

\section{Positional encoding}
\begin{figure*}[h!]
\includegraphics[width=1.0\linewidth]{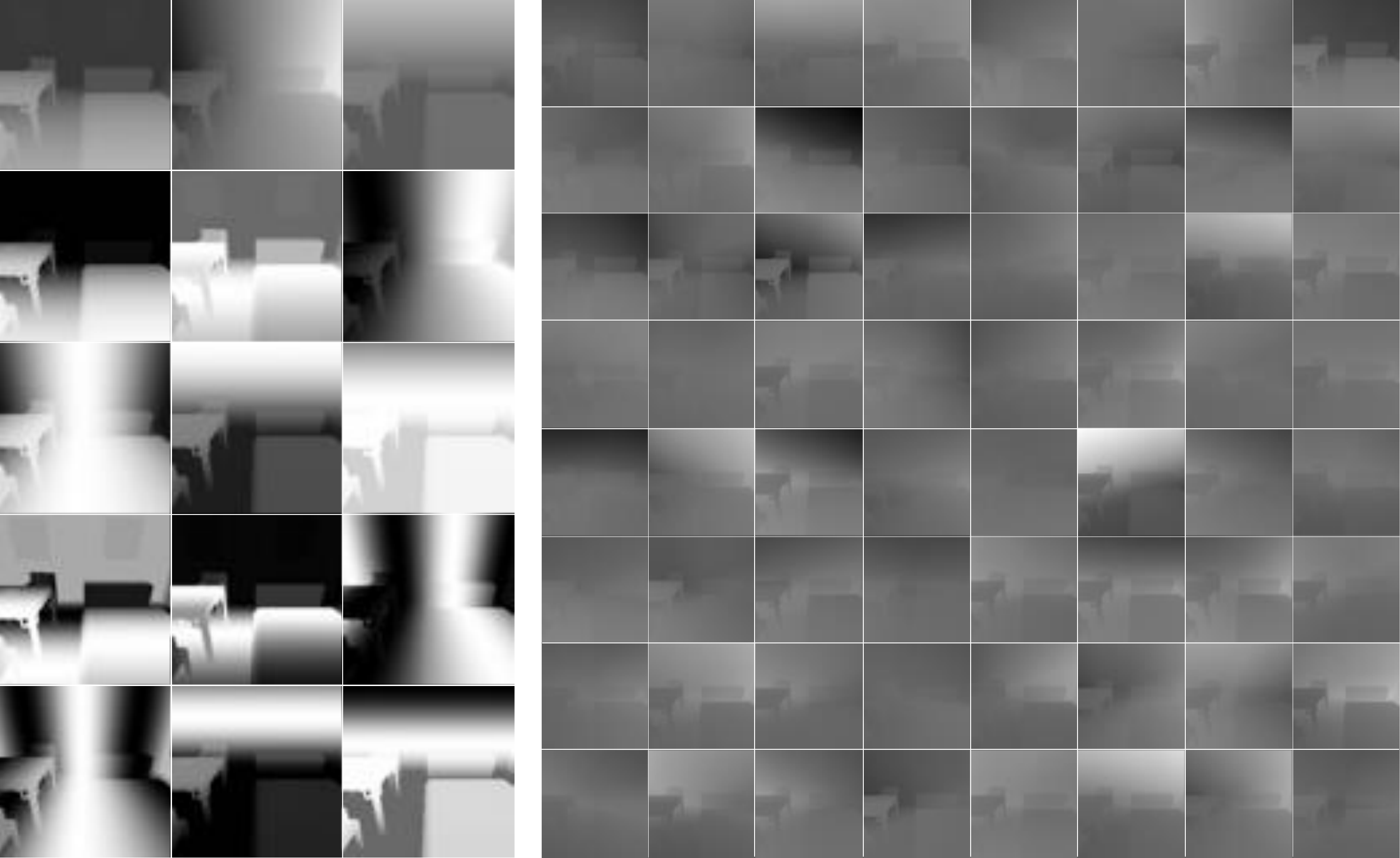}
\caption{Visualization of the positional encoding. Left: $15$ feature maps of $\gamma_{pe}$. Right: $64$ feature maps of $\gamma_{ne}$.} 
\label{fig:pos_encoding_fmap}
\end{figure*}

The positional encodings $\gamma_{pe}$ and $\gamma_{ne}$ are one of the essential components in our framework that enables the effective synthesis of natural-looking images.
To investigate the effect of each positional encoding, we visualize them in Figure~\ref{fig:pos_encoding_fmap}.
As shown in the figure, the encoding $\gamma_{pe}$ shows sinusoidal patterns with different frequencies. Increasing the dimension of $\gamma_{pe}$ introduces high-frequency sinusoidal patterns, which eventually lead to high-frequency artifacts, as reported in our main paper.
To resolve this, we adopt the learnable encoding $\gamma_{ne}$. As shown in the figure, the encoding $\gamma_{ne}$ does not show noticeable repeating patterns while still providing the position information.
As a result, with $\gamma_{ne}$, we can avoid high-frequency artifacts caused by high-dimensional $\gamma_{pe}$.

\begin{figure*}[t!]
\centering
\includegraphics[width=0.85\linewidth]{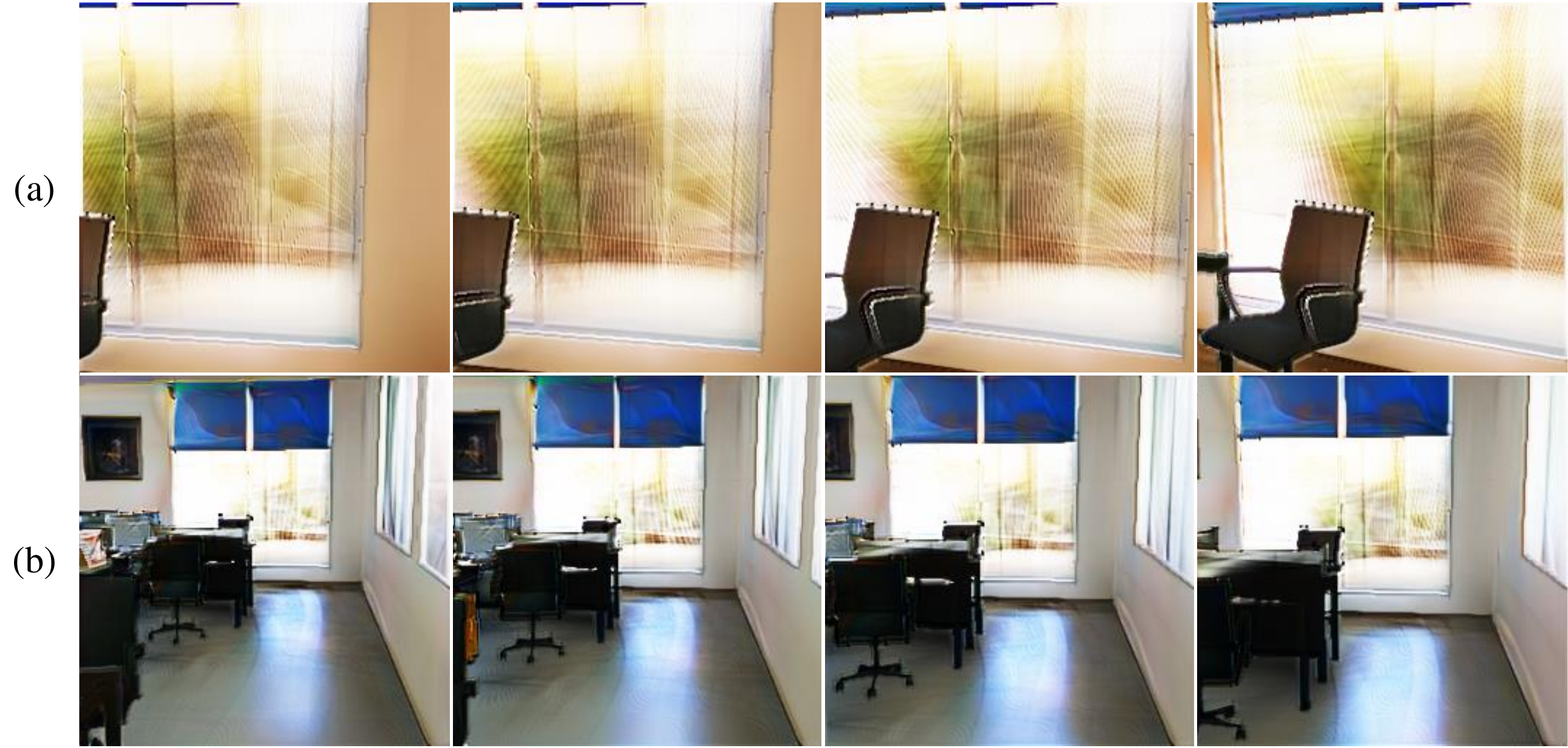}
\caption{Failure cases. (a) Our method cannot handle transparent objects such as windows. (b) Our method cannot effectively model view-dependent components such as specular lighting and reflection.} 
\label{fig:failure}
\end{figure*}

\begin{figure*}[t!]
\centering
\includegraphics[width=0.85\linewidth]{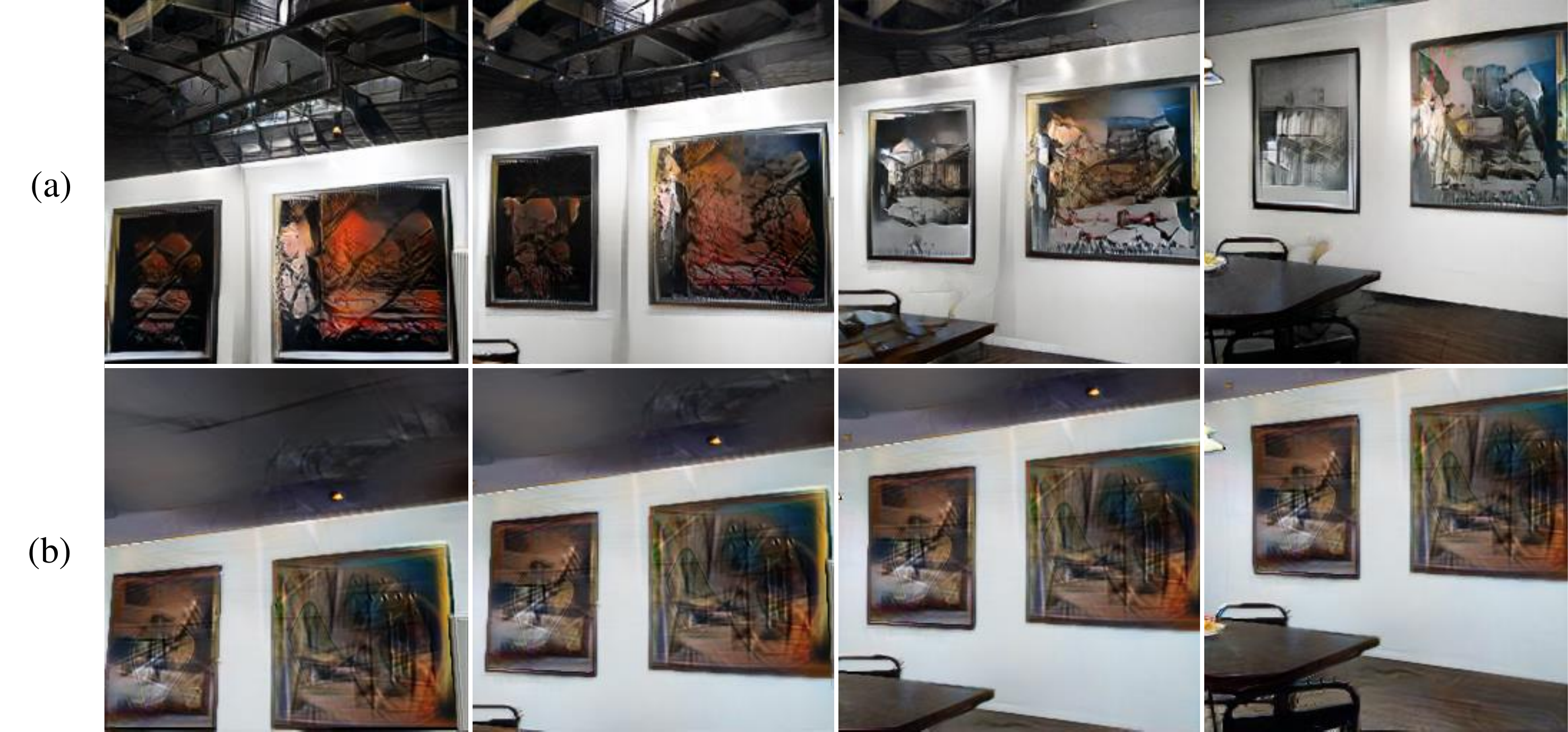}
\caption{Failure cases. Our method aggregates independently synthesized images. This may introduce blurry artifacts.  (a) Reference images generated by OASIS \cite{Sushko2020YouON}. The ceiling has complex textures (b) Our results have ceilings with smooth textures.} 
\label{fig:texture}
\end{figure*}

\section{Failure Cases}
Our framework is based on the learning of a mapping from a 3D coordinate to a color value.
This approach introduces a few limitations.
First, our method cannot handle transparent objects such as windows but handles them as opaque objects, as shown in \Fig{failure} (a). 
Second, our method cannot effectively handle view-dependent components in natural images such as specular lighting and reflection (\Fig{failure} (b)).
Our framework aggregates images independently synthesized by a 2D semantic image synthesis method.
This may introduce blurry artifacts.
In \Fig{texture}, while the ceiling in the reference images generated by OASIS~\cite{Sushko2020YouON} has complex textures,
the ceiling in our results looks smooth without textures.